\documentclass[11pt,a4paper]{article}
\usepackage[hyperref]{acl2018}
\usepackage{times}
\usepackage{latexsym}
\usepackage{algorithm}
\usepackage{algorithmic}
\usepackage{booktabs}
\usepackage{array}
\usepackage{color}
\usepackage{amsmath,amsfonts,amssymb,bbm,epsfig,bm}
\usepackage{float}
\usepackage{multirow}
\usepackage{balance}
\usepackage{url}
\usepackage{subfigure}
\newcommand{\seq}[1]{\mathbf{#1}}

\usepackage{CJK}

\newcolumntype{L}[1]{>{\raggedright\arraybackslash}p{#1}}
\newcolumntype{R}[1]{>{\raggedleft\arraybackslash}p{#1}}
\newcolumntype{?}{!{\vrule width 1pt}}

\aclfinalcopy 

\title{Automatic Article Commenting: the Task and Dataset}

\author{Lianhui Qin$^1$\thanks{Work done while Lianhui interned at Tencent AI Lab},~~
Lemao Liu$^2$,~~ 
Victoria Bi$^2$,~~ 
Yan Wang$^2$,~~\\
{\bf Xiaojiang Liu$^2$,}~~
{\bf Zhiting Hu,}~~ 
{\bf Hai Zhao$^1$,}~~ 
{\bf Shuming Shi$^2$}\\
{\small Department of Computer Science and Engineering, Shanghai Jiao Tong University$^1$, Tencent AI Lab$^2$,} \\
{\small{\tt \{lianhuiqin9,zhitinghu\}@gmail.com},~{\tt zhaohai@cs.sjtu.edu.cn},} \\ 
{\small {\tt \{victoriabi,brandenwang,lmliu,kieranliu,shumingshi\}@tencent.com}}}

\date{}

\begin{document}
\maketitle

\begin{abstract}
Comments of online articles provide extended views and improve user engagement. Automatically making comments thus become a valuable functionality for online forums, intelligent chatbots, etc. This paper proposes the new task of automatic article commenting, and introduces a large-scale Chinese dataset\footnote{The dataset is available on \url{http://ai.tencent.com/upload/PapersUploads/article_commenting.tgz}} with millions of real comments and a human-annotated subset characterizing the comments' varying quality. Incorporating the human bias of comment quality, we further develop automatic metrics that generalize a broad set of popular reference-based metrics and exhibit greatly improved correlations with human evaluations.
\end{abstract}

\section{Introduction}
Comments of online articles and posts provide extended information and rich personal views, which could attract reader attentions and improve interactions between readers and authors~\cite{park2016supporting}. In contrast, posts failing to receive comments can easily go unattended and buried. With the prevalence of online posting, automatic article commenting thus becomes a highly desirable tool for online discussion forums and social media platforms to increase user engagement and foster online communities. Besides, commenting on articles is one of the increasingly demanded skills of intelligent chatbot~\cite{shum2018eliza} to enable in-depth, content-rich conversations with humans.

\begin{table*}[t]
\vspace{-5pt}
\begin{minipage}{.4\linewidth}
\small
	\centering
	\begin{tabular}{@{} L{6cm} @{}}
	\cmidrule[\heavyrulewidth]{1-1}
	 {\bf Title:} \begin{CJK*}{UTF8}{gkai}苹果公司iPhone 8 发布会定在9月举行 \end{CJK*} (Apple's iPhone 8 event is happening in Sept.)
 \\
 	\cmidrule[\heavyrulewidth]{1-1}
    {\bf Content:} \begin{CJK*}{UTF8}{gkai}
			苹果公司正式向媒体发布邀请函，宣布将于9月12日召开苹果新品发布会，该公司将发布下一代iPhone， 随之更新的还有苹果手表，苹果TV， 和iOS软件。
			这次发布会将带来三款新iPhones：带OLED显示屏和3D人脸扫描技术的下一代iPhone8；是iPhone 7、iPhone 7Plus的更新版。 
		 \end{CJK*}
    
   (Apple has sent out invites for its next big event on September 12th, where the company is expected to reveal the next iPhone, along with updates to the Apple Watch, Apple TV, and iOS software.
Apple is expected to announce three new iPhones at the event: a next-generation iPhone 8 model with an OLED display and a 3D face-scanning camera; and updated versions of the iPhone 7 and 7 Plus.)
\\
	\cmidrule[\heavyrulewidth]{1-1}
	\end{tabular}
\end{minipage}
\begin{minipage}{.6\linewidth}
	\centering
\small
	\begin{tabular}{@{}R{0.4cm} L{2.9cm} L{5.3cm}@{}}
	\cmidrule[\heavyrulewidth]{1-3}
	 \bf{Score} & \bf{Criteria} & \bf{Example Comments} \\
	\cmidrule[\heavyrulewidth]{1-3}
     5 & Rich in content; attractive; deep insights; new yet relevant viewpoints & \begin{CJK*}{UTF8}{gkai}还记得那年iphone 4发布后随之而来的关于iPhone 5的传闻吗? 如果苹果今年也是这样我会觉得很滑稽。\end{CJK*} (Remember a year of iPhone 5 rumors followed by the announcement of the iPhone 4S? I will be highly entertained if Apple does something similar.)
\\ \cmidrule{1-3}
     4 & Highly relevant with meaningful ideas & \begin{CJK*}{UTF8}{gkai}就说：我们相约在那个公园。\end{CJK*}\quad(Could have said: Meet us at the Park.)
 \\ \cmidrule{1-3}
     3 & Less relevant; applied to other articles & \begin{CJK*}{UTF8}{gkai}很期待这件事！\end{CJK*}\quad\quad\quad\quad\quad\quad\quad\quad(Looking forward to this event!)
 \\ \cmidrule{1-3}
     2 & Fluent/grammatical; irrelevant &   \begin{CJK*}{UTF8}{gkai}我喜欢这只猫， 它很可爱！！\end{CJK*} \quad\quad\quad(I like the cat. it is so cute !) \\ \cmidrule{1-3}
     1 & Hard to read; Broken language; Only emoji & \begin{CJK*}{UTF8}{gkai} LOL。。。！！！\end{CJK*}\quad\quad\quad\quad\quad\quad\quad\quad(LOL... !!!) \\
	\cmidrule[\heavyrulewidth]{1-3}
	\end{tabular}
\end{minipage}
\vspace{-10pt}
    \caption{A data example of an article (including title and content) paired with selected comments.
    We also list a brief version of human judgment criteria (more details are in the supplement).}
\label{tab:example-data}
\vspace{-8pt}
\end{table*}

Article commenting poses new challenges for machines, as it involves multiple cognitive abilities: understanding the given article, 
formulating opinions and arguments, and organizing natural language for expression. Compared to summarization ~\cite{hovy1998automated}, a comment does not necessarily cover all salient ideas of the article; instead it is often desirable for a comment to carry additional information not explicitly presented in the articles. Article commenting also differs from making product reviews~\cite{Tang2017,li2017neural}, as the latter takes structured data (e.g., product attributes) as input; while the input of article commenting is in plain text format, posing a much larger input space to explore. 

In this paper, we propose the new task of automatic article commenting, and release a large-scale Chinese corpus with a human-annotated subset for scientific research and evaluation. We further develop a general approach of enhancing popular automatic metrics, such as BLEU~\cite{papineni2002bleu} and METEOR~\cite{banerjee2005meteor}, to better fit the characteristics of the new task. In recent years, enormous efforts have been made in different contexts that analyze one or more aspects of online comments. For example, \citet{kolhatkar2017constructive} identify constructive news comments; \citet{barker2016sensei} study human summaries of online comment conversations.
The datasets used in these works are typically not directly applicable in the context of article commenting, and are small in scale that is unable to support the unique complexity of the new task.

In contrast, our dataset consists of around 200K news articles and 4.5M human comments along with rich meta data for article categories and user votes of comments. Different from traditional text generation tasks such as machine translation~\cite{brown1990statistical} that has a relatively small set of gold targets, human comments on an article live in much larger space by involving diverse topics and personal views, and critically, are of varying quality in terms of readability, relevance, argument quality, informativeness, etc~\cite{diakopoulos2015picking,park2016supporting}. We thus ask human annotators to manually score a subset of over 43K comments based on carefully designed criteria for comment quality. The annotated scores reflect human's cognitive bias of comment quality in the large comment space. Incorporating the scores in a broad set of automatic evaluation metrics, we obtain enhanced metrics that exhibit greatly improved correlations with human evaluations. 
We demonstrate the use of the introduced dataset and metrics by testing on simple retrieval and seq2seq generation models. We leave more advanced modeling of the article commenting task for future research.
\begin{table}[t]
\small
	\centering
	\begin{tabular}{r l l l}
	\cmidrule[\heavyrulewidth]{1-4}
	  & Train & Dev & Test \\ 
	\cmidrule{1-4}
    \#Articles & 191,502 & 5,000 & 1,610 \\
    \#Cmts/Articles & 27 & 27 & 27 \\
    \#Upvotes/Cmt & 5.9 & 4.9 & 3.4 \\ \cmidrule{1-4}
	\cmidrule[\heavyrulewidth]{1-4}
	\end{tabular}
	\caption{Data statistics.} 
    \label{tab:stats}
\end{table}
%
\section{Related Work}
There is a surge of interest in natural language generation tasks, such as machine translation~\cite{brown1990statistical,bahdanau2014neural}, dialog~\cite{williams2007partially,shum2018eliza}, text manipulation~\cite{hu2017toward}, visual description generation~\cite{vinyals2015show,liang2017recurrent}, and so forth. Automatic article commenting poses new challenges due to the large input and output spaces and the open-domain nature of comments.

Many efforts have been devoted to studying specific attributes of reader comments, such as constructiveness, persuasiveness, and sentiment~\cite{wei2016post,kolhatkar2017constructive,barker2016sensei}. We introduce the new task of generating comments, and develop a dataset that is orders-of-magnitude larger than previous related corpus. Instead of restricting to one or few specific aspects, we focus on the general comment quality aligned with human judgment, and provide over 27 gold references for each data instance to enable wide-coverage  evaluation. Such setting also allows a large output space, and makes the task challenging and valuable for text generation research.
\citet{yao2017automated} explore defense approaches of spam or malicious reviews. We believe the proposed task and dataset can be potentially useful for the study.

\citet{galley2015deltableu} propose $\Delta$BLEU that weights multiple references for conversation generation evaluation. The quality weighted metrics developed in our work can be seen as a generalization of $\Delta$BLEU to many popular reference-based metrics (e.g., METEOR, ROUGE, and CIDEr). Our human survey demonstrates the effectiveness of the generalized metrics in the article commenting task.

\section{Article Commenting Dataset}\label{sec:dataset}
The dataset is collected from Tencent News (news.qq.com), one of the most popular Chinese websites of news and opinion articles. 
Table~\ref{tab:example-data} shows an example data instance in the dataset (For readability we also provide the English translation of the example). 
Each instance has a title and text content of the article, a set of reader comments, and side information (omitted in the example) including the article category assigned by editors, and the number of user upvotes of each comment. 

We crawled a large volume of articles posted in Apr--Aug 2017, tokenized all text with the popular python library Jieba,
and filtered out short articles with less than 30 words in content and those with less than 20 comments. The resulting corpus is split into train/dev/test sets. The selection and annotation of the test set are described shortly. 
Table~\ref{tab:stats} provides the key data statistics. The dataset has a vocabulary size of 1,858,452. The average lengths of the article titles and content are 15 and 554 Chinese words (not characters), respectively. The average comment length is 17 words.

Notably, the dataset contains an enormous volume of tokens, and is orders-of-magnitude larger than previous public data of article comment analysis~\cite{wei2016post,barker2016sensei}. Moreover, each article in the dataset has on average over 27 human-written comments. Compared to other popular text generation tasks and datasets~\cite{chen2015microsoft,wiseman2017challenges} which typically contain no more than 5 gold references, our dataset enables richer guidance for model training and wider coverage for evaluation, in order to fit the unique large output space of the commenting task.
Each article is associated with one of 44 categories, whose distribution is shown in the supplements. The number of upvotes per comment ranges from 3.4 to 5.9 on average. Though the numbers look small, the distribution exhibits a long-tail pattern with popular comments having thousands of upvotes.

\vspace{-4pt}
\paragraph{Test Set Comment Quality Annotations}
Real human comments are of varying quality. Selecting high-quality gold reference comments is necessary to encourage high-quality comment generation, and for faithful automatic evaluation, especially with reference-based metrics (sec.\ref{sec:metrics}). The upvote count of a comment is shown not to be a satisfactory indicator of its quality~\cite{park2016supporting,wei2016post}. We thus curate a subset of data instances for human annotation of comment quality, which is also used for enhancing automatic metrics as in the next section.

Specifically, we randomly select a set of 1,610 articles such that each article has at least 30 comments, each of which contains more than 5 words, and has over 200 upvotes for its comments in total. Manual inspection shows such articles and comments tend to be meaningful and receive lots of readings. We then randomly sample 27 comments for each of the articles, and ask 5 professional annotators to rate the comments. The criteria are adapted from previous journalistic criteria study~\cite{diakopoulos2015picking} and are briefed in Table~\ref{tab:example-data}, right panel (More details are provided in the supplements).   
Each comment is randomly assigned to two annotators who are presented with the criteria and several examples for each of the quality levels. The inter-annotator agreement measured by the Cohen's $\kappa$ score~\cite{cohen1968weighted} is 0.59, which indicates moderate agreement and is better or comparable to previous human studies in similar context~\cite{lowe2017towards,liu2016not}. The average human score of the test set comments is 3.6 with a standard deviation of 0.6, and 20\% of the comments received at least one 5 grade. This shows the overall quality of the test set comments is good, though variations do exist.
%
\section{Quality Weighted Automatic Metrics}\label{sec:metrics}
Automatic metrics, especially the reference-based metrics such as BLEU~\cite{papineni2002bleu}, METEOR~\cite{banerjee2005meteor}, ROUGE~\cite{lin:2004}, CIDEr~\cite{vedantam+:2015}, are widely used in text generation evaluations. These metrics have assumed all references are of equal golden qualities. However, in the task of article commenting, the real human comments as references are of varying quality as shown in the above human annotations. It is thus desirable to go beyond the equality assumption, and account for the different quality scores of the references.
This section introduces a series of enhanced metrics generalized from respective existing metrics, for leveraging human biases of reference quality and improving metric correlations with human evaluations. 

\begin{figure*}[t]
\centering
\begin{minipage}{0.3\linewidth}
\begin{center}
 \includegraphics[width=\linewidth]{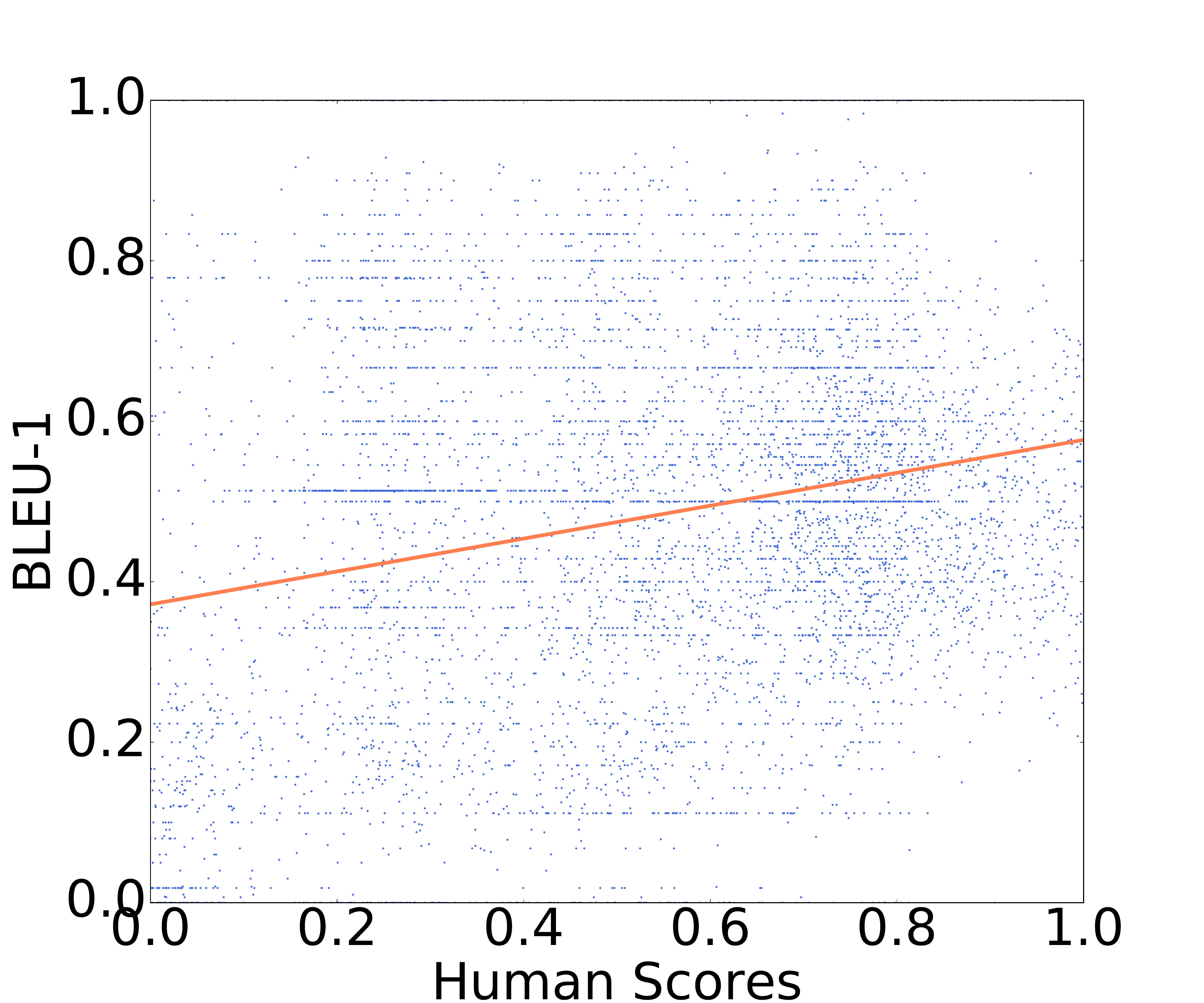}
\end{center}
\end{minipage}
\begin{minipage}{0.3\linewidth}
\begin{center}
 \includegraphics[width=\linewidth]{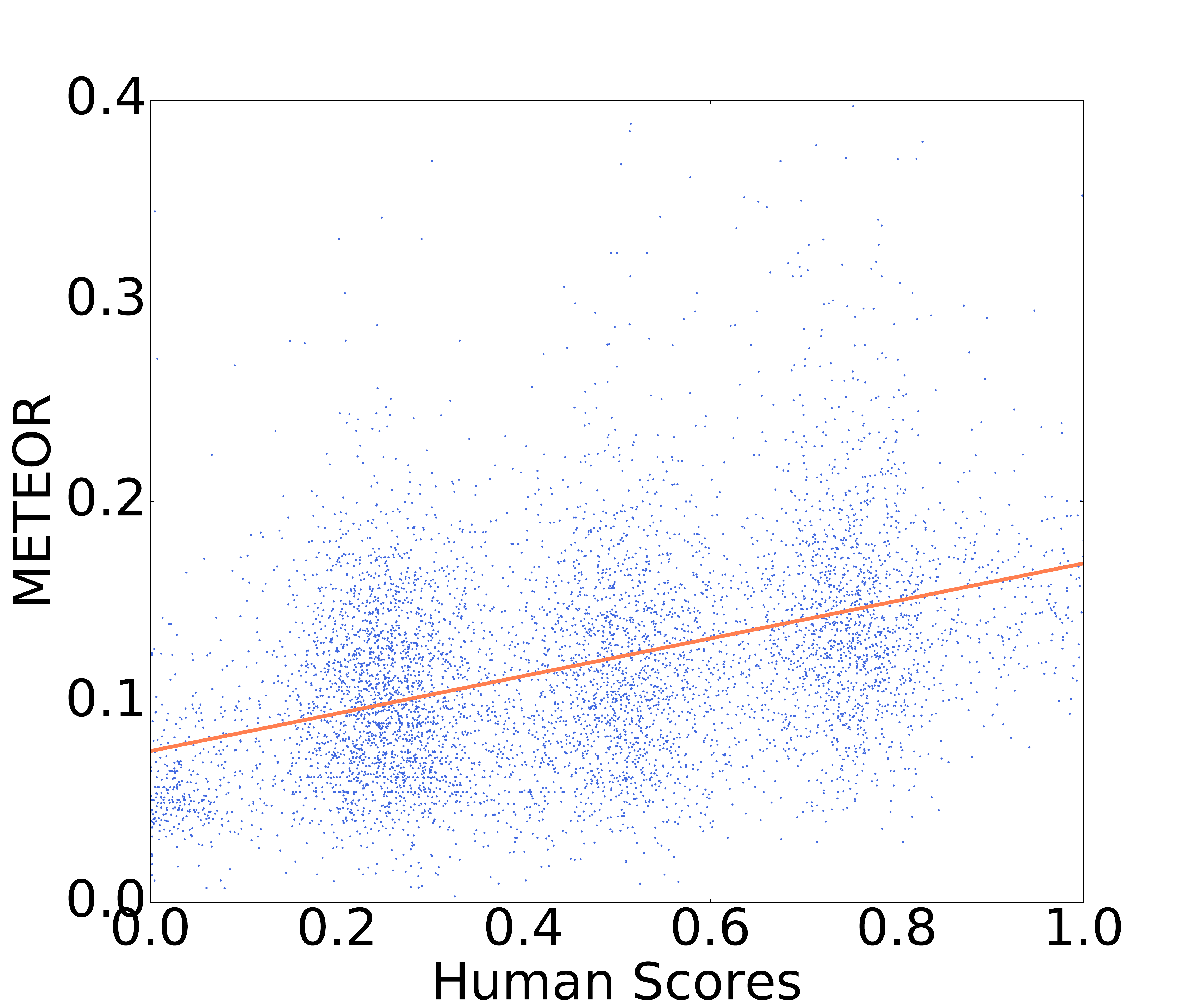}
\end{center}
\end{minipage}
\begin{minipage}{0.3\linewidth}
\begin{center}
 \includegraphics[width=\linewidth]{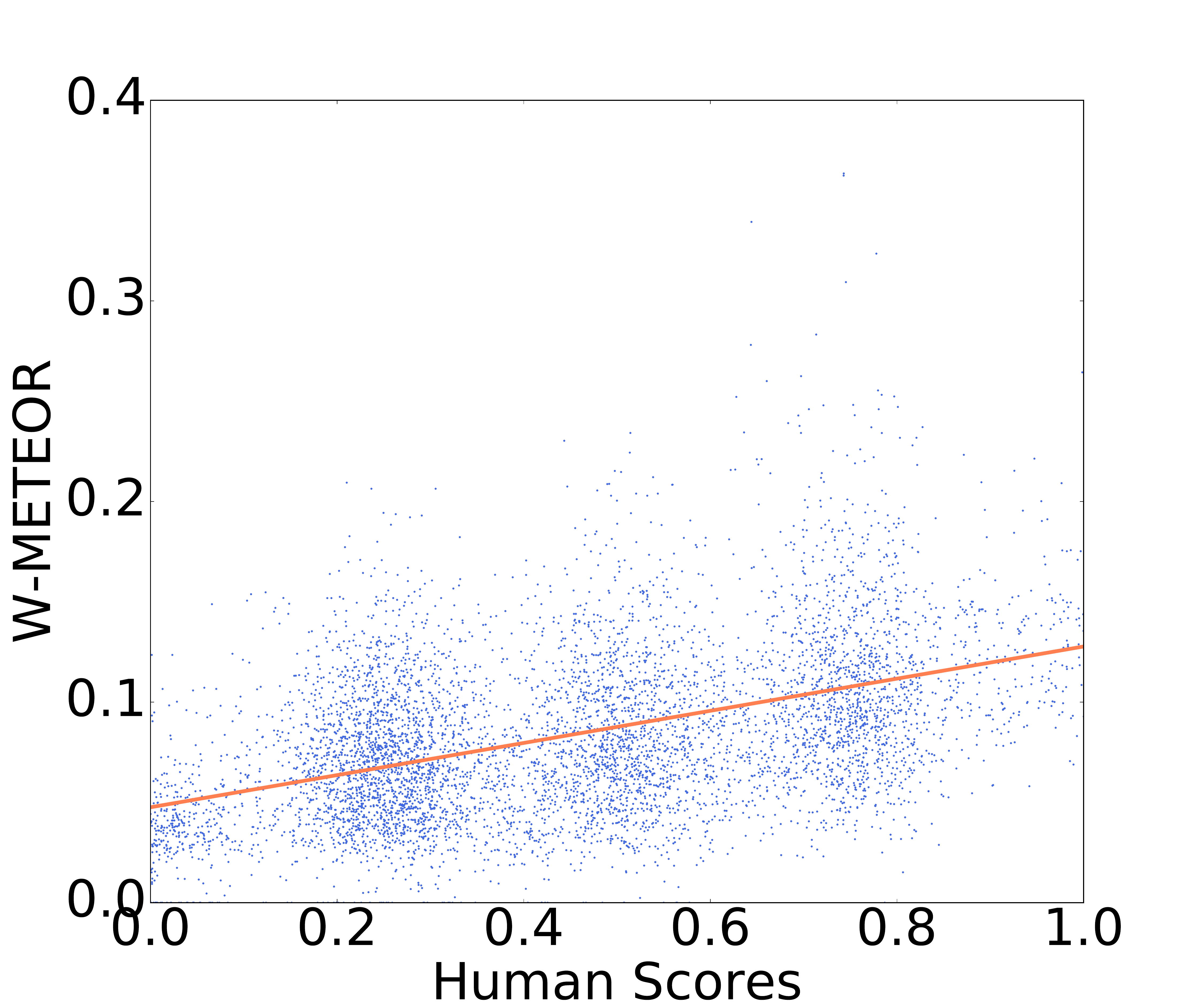}
\end{center}
\end{minipage}
\caption{Scatter plots showing the correlation between metrics and human judgments. {\bf Left: }BLEU-1; {\bf Middle: }METEOR; {\bf Right: }W-METEOR. Following~\cite{lowe2017towards}, we added Gaussian noise drawn from $\mathcal{N}(0,0.05)$ to the integer human scores to better visualize the density of points.}
\label{fig:corr}
\end{figure*}

Let $\seq{c}$ be a generated comment to evaluate, $\mathcal{R} = \{\seq{r}^j\}$ the set of references, each of which has a quality score $s^j$ by human annotators. We assume properly normalized $s^j\in [0,1]$.
Due to space limitations, here we only present the enhanced METEOR, and defer the formulations of enhancing BLEU, ROUGE, and CIDEr to the supplements. Specifically, METEOR performs word matching through an alignment between the candidate and references. The {\it weighted METEOR} extends the original metric by weighting references with $s^j$:
\begin{equation}
	\small
	\text{W-METEOR}(\seq{c}, \mathcal{R}) = (1-BP) \max\nolimits_j s^j F_{mean,j},
\end{equation}
where $F_{mean,j}$ is a harmonic mean of the precision and recall between $\seq{c}$ and $\seq{r}^j$, and $BP$
is the penalty~\cite{banerjee2005meteor}. Note that the new metrics fall back to the respective original metrics by setting $s^j=1$.

\section{Experiments}
We demonstrate the use of the dataset and metrics with simple retrieval and generation models, and show the enhanced metrics consistently improve correlations with human judgment. Note that this paper does not aim to develop solutions for the article commenting task. We leave the advanced modeling for future work. 

\begin{table}[!h]
\small
	\centering
	\begin{tabular}{r R{1.5cm} R{1.5cm}}
	\cmidrule[\heavyrulewidth]{1-3}
	\textbf{Metric} & \textbf{Spearman} & \textbf{Pearson}
\\
		\cmidrule[\heavyrulewidth]{1-3}
        METEOR
		&0.5595
		&0.5109
        \\
        W-METEOR
        &{\bf 0.5902}
		&{\bf 0.5747}
        \\
        \cmidrule{1-3}
        Rouge\_L
		&0.1948
		&0.1951
        \\
        W-Rouge\_L
		&{\bf 0.2558}
		&{\bf 0.2572}
        \\
		\cmidrule{1-3}
        CIDEr
		&0.3426
		&0.1157
        \\
        W-CIDEr
		&{\bf 0.3539}
		&{\bf 0.1261}
        \\
		\cmidrule{1-3}
        BLEU-1
		&0.2145
		&0.1790
        \\
        W-BLEU-1
		&0.2076
		&0.1604
        \\
        BLEU-4
		&0.0983
		&0.0099
        \\
        W-BLEU-4
		&{\bf 0.0998}
		&{\bf 0.0124}
        \\
        \cmidrule{1-3}
		Human
		&0.7803
		&0.7804
        \\
		\cmidrule[\heavyrulewidth]{1-3}
	\end{tabular}
    \vspace{-10pt}
	\caption{Human correlation of metrics. ``Human'' is the results from randomly dividing human scores into two groups. All p-value $<0.01$.}
    \label{tab:correlation}
    \vspace{-12pt}
\end{table}

\paragraph{Setup}
We briefly present key setup, and defer more details to the supplements. Given an article to comment, the retrieval-based models first find a set of similar articles in the training set by TF-IDF, and return the comments most relevant to the target article with a CNN-based relevance predictor. We use either the article title or full title/content for the article retrieval, and denote the two models with {\it IR-T} and {\it IR-TC}, respectively. The generation models are based on simple sequence-to-sequence network~\cite{sutskever2014sequence}. The models read articles using an encoder and generate comments using a decoder with or without attentions~\cite{bahdanau2014neural}, which are denoted as {\it Seq2seq} and {\it Att} if only article titles are read. We also set up an attentional sequence-to-sequence model that reads full article title/content, and denote with {\it Att-TC}. Again, these approaches are mainly for demonstration purpose and for evaluating the metrics, and are far from solving the difficult commenting task.
We discard comments with over 50 words and use a truncated vocabulary of size 30K. 

\paragraph{Results}
We follow previous setting~\cite{papineni2002bleu,liu2016not,lowe2017towards} to evaluate the metrics, by conducting human evaluations and calculating the correlation between the scores assigned by humans and the metrics.
Specifically, for each article in the test set, we obtained six comments, five of which come from IR-T, IR-TC, Seq2seq, Att, and Att-TC, respectively, and one randomly drawn from real comments that are different from the reference comments. The comments were then graded by human annotators following the same procedure of test set scoring (sec.\ref{sec:dataset}).
Meanwhile, we measure each comment with the vanilla and weighted automatic metrics based on the reference comments. 

Table~\ref{tab:correlation} shows the Spearman and Pearson coefficients between the comment scores assigned by humans and the metrics. The METEOR family correlates best with human judgments, and the enhanced weighted metrics improve over their vanilla versions in most cases (including BLEU-2/3 as in the supplements). E.g., the Pearson of METEOR is substantially improved from 0.51 to 0.57, and the Spearman of ROUGE\_L from 0.19 to 0.26.
Figure~\ref{fig:corr} visualizes the human correlation of BLEU-1, METEOR, and W-METEOR, showing that the BLEU-1 scores vary a lot given any fixed human score, appearing to be random noise, while the METEOR family exhibit strong consistency with human scores. Compared to W-METEOR, METEOR deviates from the regression line more frequently, esp. by assigning unexpectedly high scores to comments with low human grades. 

Notably, the best automatic metric, W-METEOR, achieves 0.59 Spearman and 0.57 Pearson, which is higher or comparable to automatic metrics in other generation tasks~\cite{lowe2017towards,liu2016not,sharma2017nlgeval,agarwal2008meteor}, indicating a good supplement to human judgment for efficient evaluation and comparison. We use the metrics to evaluate the above models in the supplements.

\section{Conclusions and Future Work}
We have introduced the new task and dataset for automatic article commenting, as well as developed quality-weighted automatic metrics that leverage valuable human bias on comment quality. The dataset and the study of metrics establish a testbed for the article commenting task.

We are excited to study solutions for the task in the future, by building advanced deep generative models~\cite{Goodfellow-et-al-2016,hu2018unifying} that incorporate effective reading comprehension modules~\cite{rajpurkar2016squad,richardson2013mctest} and rich external knowledge~\cite{angeli2015leveraging,hu2016harnessing}. 

The large dataset is also potentially useful for a variety of other tasks, such as comment ranking~\cite{hsu2009ranking}, upvotes prediction~\cite{rizos2016predicting}, and article headline generation~\cite{banko2000headline}. 
We encourage the use of the dataset in these context.

\vspace{10pt}

\noindent
{\small
{\bf Acknowledgement. }We would like to thank anonymous reviewers for their helpful suggestions and particularly the annotators for their contributions on the dataset. Hai Zhao was partially supported by National Key Research and Development Program of China (No. 2017YFB0304100), 
National Natural Science Foundation of China (No. 61672343 and No. 61733011), Key Project of National Society Science Foundation of China (No. 15-ZDA041),
The Art and Science Interdisciplinary Funds of Shanghai Jiao Tong University (No. 14JCRZ04).
}

\small
\balance
\bibliography{acl2018}
\bibliographystyle{acl_natbib}

\onecolumn

\appendix

\numberwithin{equation}{section}

\section{Dataset}
\begin{figure}[!hp]
\begin{center}
 \includegraphics[width=0.5\linewidth]{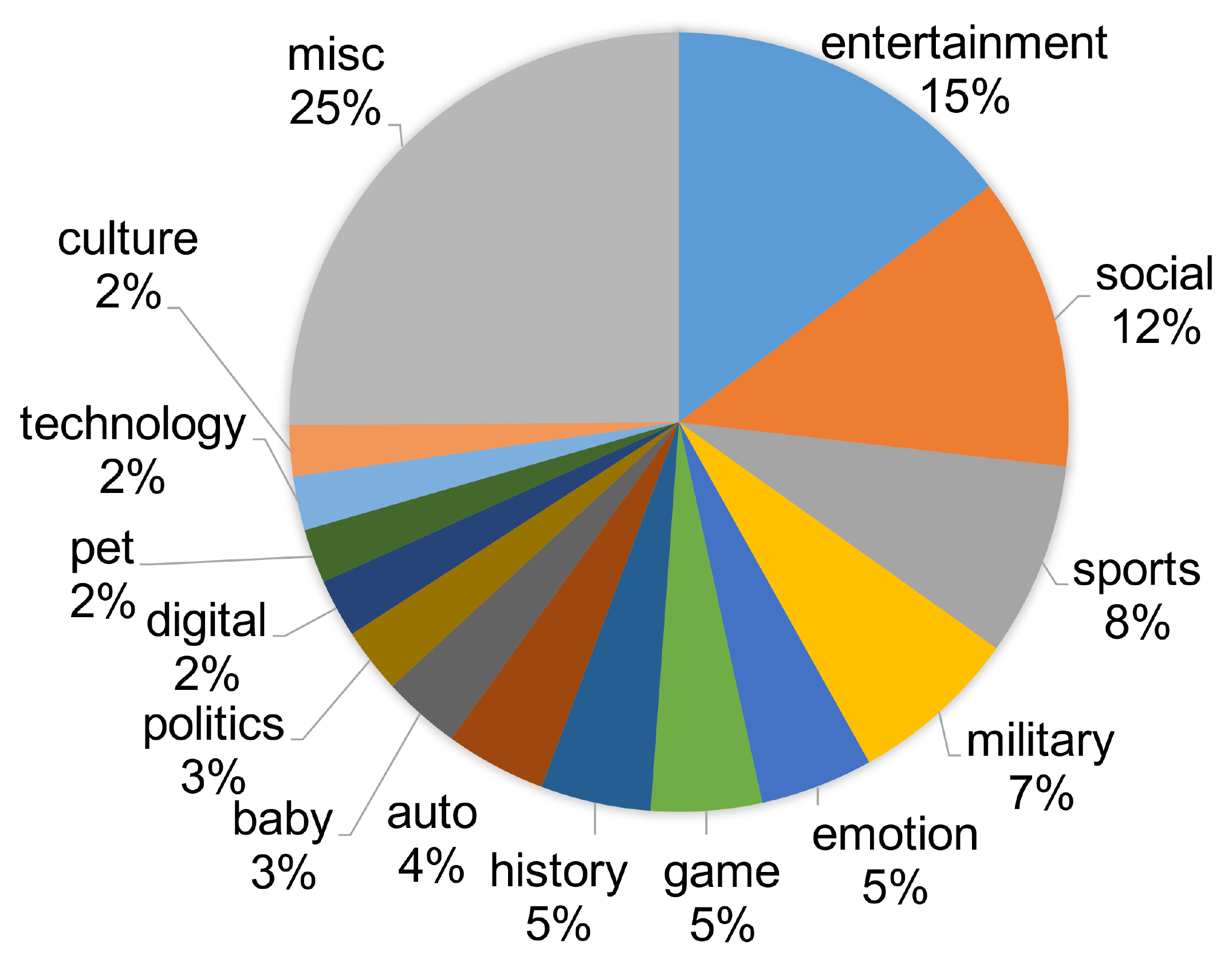}
\end{center}
\vspace{-5pt}
\caption{Category distribution of the articles in the dataset. Top 15 most frequent categories are shown.}
\label{fig:data-dist}
\end{figure}

\subsection{Human Evaluation Criteria}  
We adapt the previous journalistic criteria study~\citep{diakopoulos2015picking,park2016supporting} and setup the following evaluation criteria of comment quality:
\begin{itemize}
\item Score 1: The comment is hard to read or even is not a normal, well-formed sentence, such as messy code, meaningless words, or merely punctuation or emoji. 

\item Score 2: The language is fluent and grammatical, but the topic or argument of the comment is irrelevant to the article. Sometimes the comment relates to advertisement or spam.

\item Score 3: The comment is highly readable, and is relevant to the article to some extent. However, the topic of the comment is vague, lacking specific details or clear focus, and can be commonly applied to other articles about different stuffs.

\item Score 4: The comment is specifically relevant to the article, expresses meaningful opinions and perspectives. The idea in the comment can be common, not necessarily novel. The language is of high quality.

\item Score 5: The comment is informative, rich in content, and expresses novel, interesting, insightful personal views that are attractive to readers, and are highly relevant to the article, or extend the original perspective in the article. 
\end{itemize}

\section{Enhanced Automatic Metrics}
Most previous literatures have used automatic evaluation metrics for evaluating generation performance, especially overlapping-based metrics that determine the quality of a candidate by measuring the token overlapping between the candidate and a set of gold references. The widely-used ones of such evaluation metrics include BLEU~\citep{papineni2002bleu},
METEOR~\citep{banerjee2005meteor}, ROUGE~\citep{lin:2004}, CIDEr~\citep{vedantam+:2015}, and so forth.
These metrics have assumed that all references are with equal golden qualities.
However, in our context,
the references (collected reader comments) are of different qualities according to the above human annotation (see the dataset section).
It is thus desirable to go beyond the oversimplified assumption of equality, and take into account the different quality scores of the references.
This section introduces a series of enhanced metrics generalized from the respective existing metrics for our specific scenario.

Suppose $\seq{c}$ is the output comment from a method, $\mathcal{R} = \{\seq{r}^1, \seq{r}^2, \cdots, \seq{r}^{K}\}$ is a set of $K$
reference comments, 
each of which has a score $s^j$ rated by human annotators indicating the quality of the reference comment. 
We assume each $s^j$ is properly normalized so that $s^j\in [0,1]$.
In the rest of the section, we describe the definitions of our enhanced metrics with weights $s^j$. Each of the new metrics falls back to the respective original metric by setting $s^j=1$. 

\subsection{Weighted BLEU}
Similarly to BLEU~\citep{papineni2002bleu}, our weighted BLEU is based on a modified precision of $n$-grams in $\seq{c}$ with respect to $\mathcal{R}$ as follows:
\begin{equation}
	\small
	\text{W-BLEU\_}N(\seq{c}, \mathcal{R}) = BP \cdot \exp(\sum_{n=1}^{N} \frac{1}{N} \log {PRC}_n),
	\label{eq:bleu}
\end{equation}
where $N$ is the maximal length of grams considered; 
$BP$ is a penalty discouraging short generations. Here we omit the definition of $BP$ due to the space limitations and refer readers to \citep{papineni2002bleu}.
Besides, ${PRC}_n$ in Eq.\eqref{eq:bleu} is the {\it weighted} precision of all $n$-grams in $\seq{c}$ regarding to $\mathcal{R}$, which is defined as follows:
\begin{equation}
	\small
	{PRC}_n=
    \frac{\sum_{\omega_n} \min\big\{\text{Count}(\omega_n, \seq{c}), \max_j s^j\text{Count}(\omega_n, \seq{r}^j)\big\} } {\sum_{\omega_n}  \text{Count}(\omega_n, \seq{c})},
\end{equation}
where $\text{Count}(\omega_n, \seq{c})$ denotes the number of times an $n$-gram $\omega_n$ occurring in $\seq{c}$. Note that each $\text{Count}(\omega_n, \seq{r}^j)$ is weighted by the score $s^j$ of reference $\seq{r}^j$. By weighting with $s^j$, overlapping with an $n$-gram of reference $\seq{r}^j$ yields a contribution proportional to the respective reference score.

\subsection{Weighted METEOR}
METEOR~\citep{banerjee2005meteor} explicitly performs word matching through an one-to-one alignment between the candidate and reference. 
Similar to METEOR, weighted METEOR requires both precision and recall based on the alignment:
the precision is defined as the ratio between the number of aligned words and the total
number of words in $\seq{c}$, and the recall is defined as the ratio between 
the number of aligned words and the total of words in $\seq{r}^j$.
The weighted METEOR is obtained by weighting reference with $s^j$ as:
\begin{equation}
	\small
	\text{W-METEOR}(\seq{c}, \mathcal{R}) = (1-BP) \max\nolimits_j s^j F_{mean,j},
\end{equation}
where $F_{mean,j}$ is a harmonic mean of the precision and recall between $\seq{c}$ and $\seq{r}^j$, and $BP$
is the penalty as defined in original METEOR~\citep{banerjee2005meteor}. 

\subsection{Weighted ROUGE}
Unlike BLEU, ROUGE biases to recall rather than precision. 
ROUGE has different implementations, and we use ROUGE-L in our experiments following \citep{liu2016not}. 
Weighted ROUGE-L is based on the longest common
subsequence (LCS) between candidate $\seq{c}$ and reference set $\mathcal{R}$:
\begin{equation}
	\small
	\text{W-ROUGE-L}(\seq{c}, \mathcal{R}) = \frac{(1+\beta^2){PRC}\times{REC}}{{REC}+\beta^2\times{PRC}},
\end{equation}
where $\beta$ is a predefined constant, and $PRC$ and $REC$ are {\it weighted} precision and recall, respectively, defined as:
\begin{equation*}
	\begin{split}
		\small
		&PRC = \frac{|\cup_j s^j LCS(\seq{c}, \seq{r}^j)|}{|\seq{c}|}, \\
        &REC = \frac{|\cup_j s^j LCS(\seq{c}, \seq{r}^j)|}{|\seq{r}^j|}.
	\end{split}
\end{equation*}
Here $LCS$ is the longest common subsequence over
a pair of sequences; $|\cup_j s^j A_j|$ denotes the length of the union of multiple sets $\{A_j\}$~\citep{lin:2004} where each set $A_j$ is weighted by $s^j$. By associating weight $s^j$ to the tokens in $LCS(\seq{c}, \seq{r}^j)$, 
each token contributes proportional to the respective weight when computing the length of union LCS.

\subsection{Weighted CIDEr}
CIDEr is a consensus-based evaluation metric that is
originally used in image description tasks.
The weighted CIDEr is defined by weighting each reference $\seq{r}_j$ with $s^j$ as follows:
\begin{equation}
	\small
	\text{W-CIDEr}(\seq{c}, \mathcal{R}) = \frac{1}{K}\sum\nolimits_n \beta_n \sum\nolimits_j s^j \cos({\mathbf g}^n(\seq{c}),{\mathbf g}^n(\seq{r}^j)),
	\label{eq:cider}
\end{equation}
where $\beta_n$ is typically set to $1/N$ with $N$ the highest order of grams; $\mathbf{g}^n(\seq{c})$ denotes the TF-IDF vector of the $n$-grams in $\seq{c}$. Note that cosine similarity with respect to each $\bm{r}_j$ is weighted by $s_j$.

Note that though the above metrics are defined for one comment at sentence level, they can be straightforwardly extended to many comments at the corpus level by aggregating respective statistics as with the original un-weighted metrics~\citep{papineni2002bleu,banerjee2005meteor}.

\section{Experiment}
\subsection{Setup}
Following the standard preprocessing steps~\citep{Britz:2017}, we truncated all comments to have maximal length of 50 words, kept 30K most frequent words in the vocabulary, and replaced infrequent ones with a special $<$unk$>$ token. The models were then trained on the pre-processed (article, comment) pairs. Note that an article can appear in multiple training pairs (We also tried randomly sampling only one comment for each title as training data, but obtained inferior model performance).
Key hyperparameters were tuned on the development set. In particular, all Seq2seq models have hidden size of 256, and were trained with Adam stochastic gradient descent~\citep{kingma2014adam}.

The basic idea of retrieval models is to find a comment $\seq{c}$ from the training data that best matches the content of article $\seq{x}$ according to a relevance model. Our retrieval models involve two stages:
(1) Retrieve a set of candidate articles for $\seq{x}$ under some similarity metrics;
(2) Set the candidate comments as the union of all comments from each retrieved article and return the best comment $\seq{c}$ according to a relevance model between $\seq{x}$ and a candidate comment. In the first stage, we employ the TF-IDF vector to retrieve a set of candidate articles according to the following metric: 
\begin{equation}
\cos(\seq{x},\seq{y}) = \cos\big(\mathbf{g}(\seq{x}), \mathbf{g}(\seq{y})\big),
\label{eq:vs}
\end{equation}
where $\mathbf{g}(\seq{x})$ is the TF-IDF weighted vector regarding to all uni-gram in $\seq{x}$~\citep{salton+:1974}.
Suppose one retrieves a set of candidate articles $\mathcal{Y}=\big\{\seq{y}^j \mid j \in \left\{1,\dots,|\mathcal{Y}|\right\}\big\}$ for $\seq{x}$ according to Eq.\eqref{eq:vs}, and the union of comments with respect to $\mathcal{Y}$ is denoted by
$\mathcal{C}=\big\{\seq{c}^j \mid j \in \left\{1,\dots,|\mathcal{C}|\right\}\big\}$.
In the second stage, to find the best comment in $\mathcal{C}$, we use a convolutional network (CNN) that takes the article $\seq{x}$ and a comment $\seq{c}\in\mathcal{C}$ as inputs, and outputs a relevance score:
\begin{equation}
P(\seq{c}|\seq{x};\theta) = \frac{\exp\big(conv(\seq{x},\seq{c};\theta)\big)}{\sum_{\seq{c'}} \exp\big(conv(\seq{x},\seq{c'}; \theta)\big)},
\label{eq:conv}
\end{equation}
where $conv(\seq{x},\seq{c};\theta)$ denotes the CNN output value (i.e., the relevance score).
Eq.\eqref{eq:conv} involves parameter $\theta$ which needs to be trained. The positive instances for training $\theta$ are the (article, comment) pairs in the training set of the proposed data. As negative instances are not directly available, we use the negative sampling technique~\cite{mikolov+:2013} to estimate the normalization term in Eq.\eqref{eq:conv}.

\subsection{Human Correlation of Automatic Metrics}

\begin{table}[!h]
	\centering
    \small
	\begin{tabular}{R{2.5cm} R{2.5cm} R{2.5cm}}
	\cmidrule[\heavyrulewidth]{1-3}
	Metric & \textbf{Spearman} & \textbf{Pearson}
\\
		\cmidrule[\heavyrulewidth]{1-3}
        METEOR
		&0.5595
		&0.5109
        \\
        W-METEOR
        &{\bf 0.5902}
		&{\bf 0.5747}
        \\
        \cmidrule{1-3}
        Rouge\_L
		&0.1948
		&0.1951
        \\
        W-Rouge\_L
		&{\bf 0.2558}
		&{\bf 0.2572}
        \\
		\cmidrule{1-3}
        CIDEr
		&0.3426
		&0.1157
        \\
        W-CIDEr
		&{\bf 0.3539}
		&{\bf 0.1261}
        \\
		\cmidrule{1-3}
        BLEU-1
		&0.2145
		&0.1790
        \\
        W-BLEU-1
		&0.2076
		&0.1604
        \\
        BLEU-2
		&0.2224
		&0.0758
        \\
        W-BLEU-2
		&{\bf 0.2255}
		&{\bf 0.0778}
        \\
        BLEU-3
		&0.1868
		&0.0150
        \\
        W-BLEU-3
		&{\bf 0.1882}
		&{\bf 0.0203}
        \\
        BLEU-4
		&0.0983
		&0.0099
        \\
        W-BLEU-4
		&{\bf 0.0998}
		&{\bf 0.0124}
        \\
        \cmidrule{1-3}
		Human
		&0.7803
		&0.7804
        \\
		\cmidrule[\heavyrulewidth]{1-3}
	\end{tabular}
	\caption{Correlation between metrics and human judgments on comments. ``Human'' represents the results from randomly dividing human judgments into two groups. All values are with p-value $<0.01$.}
    \label{tab:correlation}
\end{table}

Table~\ref{tab:correlation} also shows consistent improvement of the weight-enhanced metrics over their vanilla versions. For instance, our proposed weighted metrics substantially improve the Pearson correlation of METEOR from 0.51 to 0.57, and the Spearman correlation of ROUGE\_L from 0.19 to 0.26.

\begin{table}[!h]
\small
	\centering
	\begin{tabular}{l L{8cm}@{}}
	\cmidrule[\heavyrulewidth]{1-2}
	{\bf Title} & \begin{CJK*}{UTF8}{gkai}徐：演技非常好的新星\end{CJK*} (Gloss: Xu: A rising star with great acting skill) \\ \cmidrule{1-2}
    {\bf Comment} & \begin{CJK*}{UTF8}{gkai}我看过她的电影《最遥远的距离》。一个充满能量和演技的演员。祝福她！\end{CJK*} (Gloss: I watched her film ``The Most Distant Course''. An actor full of power and with experienced skills. Best wishes!) \\ \cmidrule{1-2}
    \multirow{3}{*}{{\bf Scores}} & Human: {\bf 4} \\ & Normalized-METEOR: {\bf 4.2} (METEOR: 0.47) \\ & Normalized-BLEU-1: {\bf 2.7} (BLEU-1: 0.38) \\
	\cmidrule[\heavyrulewidth]{1-2}
    {\bf Title} & \begin{CJK*}{UTF8}{gkai}一张褪色的照片帮助解决了18年前的谋杀案\end{CJK*} (Gloss: A faded photo helped solve a murder that happened 18 years ago) \\ \cmidrule{1-2}
    {\bf Comment} & \begin{CJK*}{UTF8}{gkai}把他关进监狱。\end{CJK*} (Gloss: Put him in prison.) \\ \cmidrule{1-2}
    \multirow{3}{*}{{\bf Scores}} & Human: {\bf 3} \\ & Normalized-METEOR: {\bf 2.7} (METEOR: 0.1) \\ & Normalized-BLEU-1: {\bf 4.5} (BLEU-1: 0.83) \\
	\cmidrule[\heavyrulewidth]{1-2}
	\end{tabular}
	\caption{Examples showing different metric scores. For comparison between metrics, we show normalized METEOR and BLEU-1 scores (highlighted) which are normalization of respective metric scores to have the same mean and variance with human scores, and clipped to be within $[1, 5]$~\citep{lowe2017towards}. The scores in parentheses are original metric scores without normalization. Note that score without normalization are not comparable. {\bf Top: }Human and METEOR gave high scores while BLEU-1 gave a low score. {\bf Bottom: }Human and METEOR gave low scores while BLEU-1 gave a high score.}
    \label{tab:score-examples}
\end{table}

Table~\ref{tab:score-examples} presents two representative examples where METEOR and BLEU-1 gave significantly different scores. Note that for inter-metric comparison of the scores, we have normalized all metrics to have the same mean and variance with the human scores. 
In the first case, the comment has rich content. Both the human annotators and METEOR graded the comment highly. However, BLEU-1 gave a low score because the comment is long and led to a low precision. The second example illustrates a converse case.

\begin{table}[!h]
\vspace{3pt}
\small
	\centering
	\begin{tabular}{r L{8cm}@{}}
	\cmidrule[\heavyrulewidth]{1-2}
	{\bf Title} & \begin{CJK*}{UTF8}{gkai}Baby重回《跑男》\end{CJK*} (Gloss: AngelaBaby is coming back to $<$Running Man$>$) \\ \cmidrule{1-2}
    {\bf Comment} & \begin{CJK*}{UTF8}{gkai}Baby, Baby, 我爱你。\end{CJK*} (Gloss:Baby, Baby, I love you.) \\ \cmidrule{1-2} 
    \multirow{3}{*}{{\bf Scores}} & Human: {\bf 3} \\ & Normalized-METEOR: {\bf 4.8} (METEOR: 0.62) \\ & Normalized-W-METEOR: {\bf 3.8} (W-METEOR: 0.34) \\
	\cmidrule[\heavyrulewidth]{1-2}
    {\bf Title} &\begin{CJK*}{UTF8}{gkai}三兄弟在车祸中受伤。\end{CJK*} (Gloss: Three siblings injured in car crash.) \\ \cmidrule{1-2}
    {\bf Comment} & \begin{CJK*}{UTF8}{gkai}祝愿三兄弟无恙。\end{CJK*} (Gloss:I hope all is well for the three guys.) \\ \cmidrule{1-2}
    \multirow{3}{*}{{\bf Scores}} & Human: {\bf 3} \\ & Normalized-METEOR: {\bf 3.9} (METEOR: 0.40) \\ & Normalized-W-METEOR: {\bf 3.2} (W-METEOR: 0.19) \\
	\cmidrule[\heavyrulewidth]{1-2}
	\end{tabular}
	\caption{Examples showing different scores of METEOR and W-METEOR. As in Table~\ref{tab:score-examples}, for comparison across metrics, we also show normalized (W-)METEOR scores.} 
    \label{tab:score-meteor-examples}
\end{table}

Table~\ref{tab:score-meteor-examples} provides examples of (W-)METEOR scores. The comments, though relevant to the articles as they refer to the keywords (i.e., actress name ``{\it Baby}'' and the injured ``{\it three guys}''), do not contain much meaningful information. However, the vanilla METEOR metric assigns high scores because the comments overlap well with one of the gold references. W-METEOR alleviates the issue as it additionally weights the references with their human grades, and successfully downplays the effect of matching with low-quality references. We see that compared to the vanilla METEOR scores, the W-METEOR scores get closer to human judgments. The results strongly validate our intuition that differentiating the qualities of gold references and emphasizing on high-quality ones bring about great benefits.

\clearpage

\subsection{Results}
\begin{table}
	\centering
	\begin{tabular}{r r r r r r}
	\cmidrule[\heavyrulewidth]{1-6}
	Metrics& IR-T & IR-TC & Seq2seq & Att & Att-TC \\ \cmidrule[\heavyrulewidth]{1-6}
	METEOR & 0.137 & {\bf 0.138} & 0.061 & 0.084 &  0.078 \\
    W-METEOR & 0.130 & {\bf 0.131} & 0.058 & 0.080 & 0.074 \\
    Rouge\_L & 0.230 & 0.229 & 0.197 & 0.232 & {\bf 0.298} \\
    W-Rouge\_L & 0.173 & 0.172 & 0.137 & 0.165 & {\bf 0.206} \\
    CIDEr & 0.007 & 0.007 & 0.006 & {\bf 0.009} & {\bf 0.009} \\
    W-CIDEr & 0.005 & {\bf 0.006} & 0.004 & {\bf 0.006} & {\bf 0.006} \\
    BLEU-1 & 0.373 & {\bf 0.374} & 0.298 & 0.368 & 0.227 \\
    W-BLEU-1 & 0.318 & 0.320 & 0.258 & {\bf 0.324} & 0.203 \\ \cmidrule{1-6}
    Human & 2.859 & {\bf 2.879} & 1.350 &  1.678 & 2.191 \\
	\cmidrule[\heavyrulewidth]{1-6}
	\end{tabular}
	\caption{Model performance under automatic metrics and human judgments.}
    \label{tab:model-results}
\end{table}

Table~\ref{tab:model-results} compares the models with various metrics. We see that IR-TC performs best under most metrics, while all methods receive human scores lower than 3.0. It is thus highly desirable to develop advanced modeling approaches to tackle the challenges in automatic article commenting.

\section{Example instance of the proposed dataset}
Examples are provided in Tables~\ref{tab:example-1} and~\ref{tab:example-2}.

\begin{table}
    \begin{tabular}{|c|L{14cm}|}
    \hline
    \bf{Title} & \begin{CJK*}{UTF8}{gkai}勇士遭首败， 杜兰特一语点出输球真因， 让全队都心碎\end{CJK*} \\ \hline
    \multirow{13}{*}{\bf{Content}}& \begin{CJK*}{UTF8}{gkai}北京时间6月10日, nba总决赛迎来了第四场比赛的较量,总比分3-0领先的勇士意欲在客场结束系列赛, 谁知骑士彻底反弹,欧文继续高效发挥, 得到40分, 詹姆斯再次得到三双31分、 10个篮板和11次助攻,勒夫也得到23分, 骑士全场投进了24个三分球, 上半场竟得到了86分, 最终在主场以137-116大胜勇士,将总比分扳成1-3, 勇士也遭遇了季后赛的首场失利。 对于本场比赛的失利, 杜兰特在赛后采访的时候表示: “我不太想对这场比赛做过多的评论, 比赛过程大家也都看到了, 有人不想让我们轻易获胜, 并且很开心我们有机会在主场夺冠。 ” 杜兰特的表达虽然很隐晦, 但是明眼人应该都能看得出这个有人是谁,那就是nba联盟和裁判。 勇士在这场比赛中打得相当被动,尤其是首节,先发五虎共领到了11次犯规,给了骑士23次罚球,使得骑士首节就砍下了48分。 在第三场比赛, 裁判就过多的干预了比赛, 好在杜兰特最后发挥神勇,逆转了比赛。本场比赛裁判仍在努力改变比赛, 最终使得骑士赢得了最后的胜利, 这恐怕也会让勇士全队球员心碎, 毕竟他们期盼着一个公平的总决赛。 下一场一场比赛将移师奥克兰,希望那是一场球员与球员的精彩对决。 \end{CJK*}
    \\ 
    \hline
    \bf{score} & \bf{comment}\\
    \hline
    3 & \begin{CJK*}{UTF8}{gkai}你去吹得了\end{CJK*}\\
    3 & \begin{CJK*}{UTF8}{gkai}几个而已，唉，这就是不懂球的玩意 \end{CJK*}\\
    4 & \begin{CJK*}{UTF8}{gkai}骑士吹了24次犯规，勇士吹了25次犯规 \end{CJK*}\\
    4 & \begin{CJK*}{UTF8}{gkai}欧文有个回场球裁判没有吹 \end{CJK*}\\
    4 & \begin{CJK*}{UTF8}{gkai}g2第一节，别说勇士，库里自己有多少罚球？别双重标准。 \end{CJK*}\\
    2 & \begin{CJK*}{UTF8}{gkai}你三岁的智商吗？ \end{CJK*}\\
    4 & \begin{CJK*}{UTF8}{gkai}太二，第一节就给了11次犯规，24分罚球，真服了，这比赛谁还敢防守，什么垃圾联盟 \end{CJK*}\\
    4 & \begin{CJK*}{UTF8}{gkai}连nba都不干净了，看来这篮球也不能看了 \end{CJK*}\\
    4 & \begin{CJK*}{UTF8}{gkai}欧文回场球都没饶还有格林对勒服的体毛犯规 \end{CJK*}\\
    3 & \begin{CJK*}{UTF8}{gkai}小编肯定是勇士球迷 \end{CJK*}\\
    3 & \begin{CJK*}{UTF8}{gkai}你这种弱智我不想多说什么，可能你眼睛瞎吧 \end{CJK*}\\
    3 & \begin{CJK*}{UTF8}{gkai}我大学是的确是篮球裁判 \end{CJK*}\\
    4 & \begin{CJK*}{UTF8}{gkai}呵呵，这回8打5终于赢了！ \end{CJK*}\\
    4 & \begin{CJK*}{UTF8}{gkai}你确定这人员配置骑士东部会垫底？这可都是詹自己选的人…. \end{CJK*}\\
    4 & \begin{CJK*}{UTF8}{gkai}那你说说为什么全场罚球勇士36个骑士31个 \end{CJK*}\\
    2 & \begin{CJK*}{UTF8}{gkai}你这说的都不合理 \end{CJK*}\\
    4 & \begin{CJK*}{UTF8}{gkai}输了就是输了，别整好像输不起是的，前几场在勇士主场骑士也遭到了同样的待遇，再有裁判是人不可能什么动作都看到 \end{CJK*}\\
    3 & \begin{CJK*}{UTF8}{gkai}你看了吗?没看别来bb，看的人都知道黑哨，你在这瞎bb? \end{CJK*}\\
    3 & \begin{CJK*}{UTF8}{gkai}真有脸说出来，你是光看比赛技术统计还是看现场直播，不要替群体来丢这个人了，哦忘了，丢人家常便饭。 \end{CJK*}\\
    4 & \begin{CJK*}{UTF8}{gkai}jr那个很明显没有违例，球快到詹姆斯手里了哨才响另外jr给詹姆斯那个传输没有回厂 \end{CJK*}\\
    4 & \begin{CJK*}{UTF8}{gkai}很正常啊，多打一场比赛联盟可以多收入几亿美刀，转播费，赞助商，球票收入，要能抢七的话肖光头绝对要笑死!这么简单的账小学生都会算，自然不会让勇士4场就解决! \end{CJK*}\\
    4 & \begin{CJK*}{UTF8}{gkai}很正常，哈登一个人一节就可以造勇士十多个 \end{CJK*}\\
    3 & \begin{CJK*}{UTF8}{gkai}的确打不过，其实有干爹呢 \end{CJK*}\\
    3 & \begin{CJK*}{UTF8}{gkai}那外国比赛，你一个外国人还看什么 \end{CJK*}\\
    3 & \begin{CJK*}{UTF8}{gkai}还有，我不是两队球迷 \end{CJK*}\\
    3 & \begin{CJK*}{UTF8}{gkai}站着不动也吹了？ \end{CJK*}\\ \hline
    \end{tabular}
    \caption{Example instance of the dataset.}
    \label{tab:example-1}
\end{table}

\begin{table}
	\begin{tabular}{|c|L{14cm}|}
		\hline
		\bf{Title} & \begin{CJK*}{UTF8}{gkai}6年前她还是杨幂小小的助理, 如今逆袭成功, 她的身价远超杨幂\end{CJK*} \\ \hline
		\multirow{13}{*}{\bf{Content}}& \begin{CJK*}{UTF8}{gkai}小编可是大幂幂的铁杆粉丝, 她参演的每部剧, 小编无一遗漏几乎全都会看完, 没办法, 谁让人家人美演技又那么棒呢, 如今的杨幂已是家喻户晓, 在她身边有个成功逆袭的助理大家却未必知晓, 说起她的名字大家可能不熟, 但提到她主演的电视大家就明白了。 她叫徐小飒, 六年还是杨幂的助理, 2009年进去娱乐圈, 曾凭借新版电视剧红楼梦中的惜春一角进去大众视野, 她的演技确实了得, 自然这也注定了她的事业也是顺风顺水。 《多情江山》中, 由徐小飒饰演的皇后索尔娜, 人物的形象被她演绎的惟妙惟肖, 就如灵魂入体一般, 虽然她饰演的是一个反面角色, 但她的演技真是无可厚非让人记忆犹新, 再加上她漂亮的脸蛋儿女神的气质, 所有的这一切都在默默的为她加分, 为她日后的事业奠定了稳固的基础。 每个人的成功都觉得偶然的, 在做助理的时候她的天分也得到过很好的展示, 而如今的她事业和演技丝毫不输于杨幂, 她是一个聪明善良的姑娘, 人们忽然喜欢她, 希望她以后的演绎事业更上一层楼上一层一层楼, 期待她有更好的作品出来。 \end{CJK*}
		\\ 
		\hline
		\bf{score} & \bf{comment}\\
		\hline
		4 & \begin{CJK*}{UTF8}{gkai}跟杨幂是没法比, 不过也不能否定人家长的还算可以吧, 将来说不定也是一线角色呢。\end{CJK*}\\
		4 & \begin{CJK*}{UTF8}{gkai}韩国终于马上调整。 就当同学。  \end{CJK*}\\
		4 & \begin{CJK*}{UTF8}{gkai}比杨幂漂亮多了。\end{CJK*}\\
		3 & \begin{CJK*}{UTF8}{gkai}很有气质！！ \end{CJK*}\\
		2 & \begin{CJK*}{UTF8}{gkai}你的脚是香的还是咋的？ \end{CJK*}\\
		5 & \begin{CJK*}{UTF8}{gkai}杨幂都有那么好吗？不觉得, 还不是全靠吹捧出来的, 别小瞧了这些后起之秀, 超过杨幂也不是不可能 \end{CJK*}\\
		2 & \begin{CJK*}{UTF8}{gkai}干啥呢？真的有哟, 你这是。 挺好, 中兴。 \end{CJK*}\\
		3 & \begin{CJK*}{UTF8}{gkai}比杨好看多了 \end{CJK*}\\
		2 & \begin{CJK*}{UTF8}{gkai}土豪, 我无话可说了。 给你刮刮挂心怀。 火车。 沈一啊, 办公室工作。 申讨的沈浪, 美女, 厦门队, 希望我写什么, 用网的吗？你好, 没好些么？我只会摸。  \end{CJK*}\\
		5 & \begin{CJK*}{UTF8}{gkai}开什么玩笑？小编你这样做娱乐新闻的？有点职业操守好吗？你说说她身价多少？怎么就超过杨幂了？杨幂现在自己的公司一部戏赚多少你知道吗？这女演员大部分观众都叫不出她名字呢！ \end{CJK*}\\
		4 & \begin{CJK*}{UTF8}{gkai}看过她参演的《遥远的距离》。\end{CJK*}\\
		3 & \begin{CJK*}{UTF8}{gkai}总是骗我们进来, 把小编吊起来打, 同意的点赞。 \end{CJK*}\\
		4 & \begin{CJK*}{UTF8}{gkai}她在《舰在亚丁湾》里演一位军嫂欧阳春, ！ \end{CJK*}\\
		3 & \begin{CJK*}{UTF8}{gkai}还是不晓得她是谁 \end{CJK*}\\
		4 & \begin{CJK*}{UTF8}{gkai}弱弱的问一句, 杨幂是谁？？ \end{CJK*}\\
		4 & \begin{CJK*}{UTF8}{gkai}看过她演的《多情江山》, 演技确实很好, 支持你, 加油！ \end{CJK*}\\
		3 & \begin{CJK*}{UTF8}{gkai}连电视名我都没听说过 \end{CJK*}\\
		3 & \begin{CJK*}{UTF8}{gkai}那只是你认为, 不自量力的东西 \end{CJK*}\\
		3 & \begin{CJK*}{UTF8}{gkai}真有脸说出来，你是光看比赛技术统计还是看现场直播，不要替群体来丢这个人了，哦忘了，丢人家常便饭。 \end{CJK*}\\
		4 & \begin{CJK*}{UTF8}{gkai}小编简直就是胡说, 什么人吖！身价还超杨幂, \end{CJK*}\\
		4 & \begin{CJK*}{UTF8}{gkai}米露也在里面演她的侄女\end{CJK*}\\
		3 & \begin{CJK*}{UTF8}{gkai}没听说过 \end{CJK*}\\
		2 & \begin{CJK*}{UTF8}{gkai}两三拿大美女, 你早找到吗？ \end{CJK*}\\
		3 & \begin{CJK*}{UTF8}{gkai}看到大家那么可劲的骂你, 我就安心了 \end{CJK*}\\
		3 & \begin{CJK*}{UTF8}{gkai}别急可能小编故意这样黑她的让大家来骂她 \end{CJK*}\\
		4 & \begin{CJK*}{UTF8}{gkai}不认识第一眼还以为是何洁 \end{CJK*}\\ \hline
	\end{tabular}
   \caption{Example instance of the dataset.}
   \label{tab:example-2}
\end{table}

\end{document}